\documentclass[10pt,twocolumn,letterpaper]{article}

\usepackage[pagenumbers]{cvpr} 

\global\long\def\vectorize#1{\mathbf{#1}}%
\global\long\def\gvt#1{\boldsymbol{#1}}%
\global\long\def\bx{\boldsymbol{x}}%
\global\long\def\bl{\boldsymbol{l}}%
\global\long\def\bz{\boldsymbol{z}}%
\global\long\def\by{\boldsymbol{y}}%
\global\long\def\bmu{\gvt{\mu}}%
\global\long\def\bpi{\boldsymbol{\pi}}%
\global\long\def\bepsilon{\boldsymbol{\epsilon}}%
\global\long\def\and{\cap}%
\global\long\def\diag{\text{diag}}%
\usepackage{amsmath,amsfonts,bm}

\def\eqref#1{equation~\ref{#1}}

\def\1{\bm{1}}

\DeclareMathAlphabet{\mathsfit}{\encodingdefault}{\sfdefault}{m}{sl}
\SetMathAlphabet{\mathsfit}{bold}{\encodingdefault}{\sfdefault}{bx}{n}

\usepackage{graphicx}
\usepackage{amsmath}
\usepackage{amssymb}
\usepackage{booktabs}

\definecolor{cvprblue}{rgb}{0.21,0.49,0.74}
\usepackage[pagebackref,breaklinks,colorlinks,allcolors=cvprblue]{hyperref}

\usepackage{natbib}  
\usepackage{caption} 
\usepackage{algorithm}
\usepackage{algorithmic}
\usepackage{xcolor}
\usepackage{colortbl} 

\usepackage{newfloat}
\usepackage{listings}
\usepackage{booktabs}
\usepackage{multirow}
\usepackage{subcaption}

\usepackage[capitalize]{cleveref}
\crefname{section}{Sec.}{Secs.}
\Crefname{section}{Section}{Sections}
\Crefname{table}{Table}{Tables}
\crefname{table}{Tab.}{Tabs.}

\def\paperID{} 
\def\confName{CVPR}
\def\confYear{2026}

\title{An Optimal Transport-driven Approach for Cultivating Latent Space in Online
Incremental Learning}

\author{Quyen Tran$^{1,}$\thanks{Quyen Tran, Hai Nguyen, and Hoang Phan contributed equally. And the work was partly done while at Qualcomm AI, Vietnam} ,
Hai Nguyen$^{2, *}$,
Quan Dao$^{1}$,
Hoang Phan$^{3, *}$,
Linh Ngo$^{4}$,
Khoat Than$^{4}$,\\
Dinh Phung$^{5}$,
Dimitris Metaxas$^{1}$,
Trung Le$^{5}$\\
\\
$^1$Rutgers University\hspace{0.3cm}$^2$Tuft University\hspace{0.3cm}$^3$New York University\hspace{0.3cm}$^4$HUST\thanks{Hanoi University of Science and Technology}\hspace{0.3cm}$^5$Monash University\\
}

\begin{document}
\maketitle
\begin{abstract}
In online incremental learning, data continuously arrives with substantial shifts in distribution, creating a significant challenge since previous samples have limited replay when learning a new task. Prior research has typically relied on either a single adaptive centroid or fixed multiple centroids to represent each class in the latent space. However, such methods struggle when class data streams are inherently multimodal and require continual centroid updates. To overcome this, we introduce an online Mixture Model learning framework grounded in Optimal Transport theory (MMOT), where centroids evolve incrementally with new data. This approach offers two main advantages: (i) it provides a more precise characterization of complex data streams, and (ii) it enables improved class similarity estimation for unseen samples during inference through MMOT-derived centroids. Furthermore, to strengthen representation learning and mitigate catastrophic forgetting, we design a Dynamic Preservation strategy that regulates the latent space and maintains class separability over time. Experimental evaluations on benchmark datasets confirm the superior effectiveness of our proposed method.
\end{abstract}    
\vspace{-3mm}
\section{Introduction}
\label{sec:intro}
\begin{figure*}[ht] 
    \centering
    \begin{subfigure}{0.45\textwidth}
        \centering
        \includegraphics[width=0.85\textwidth, height=6cm]{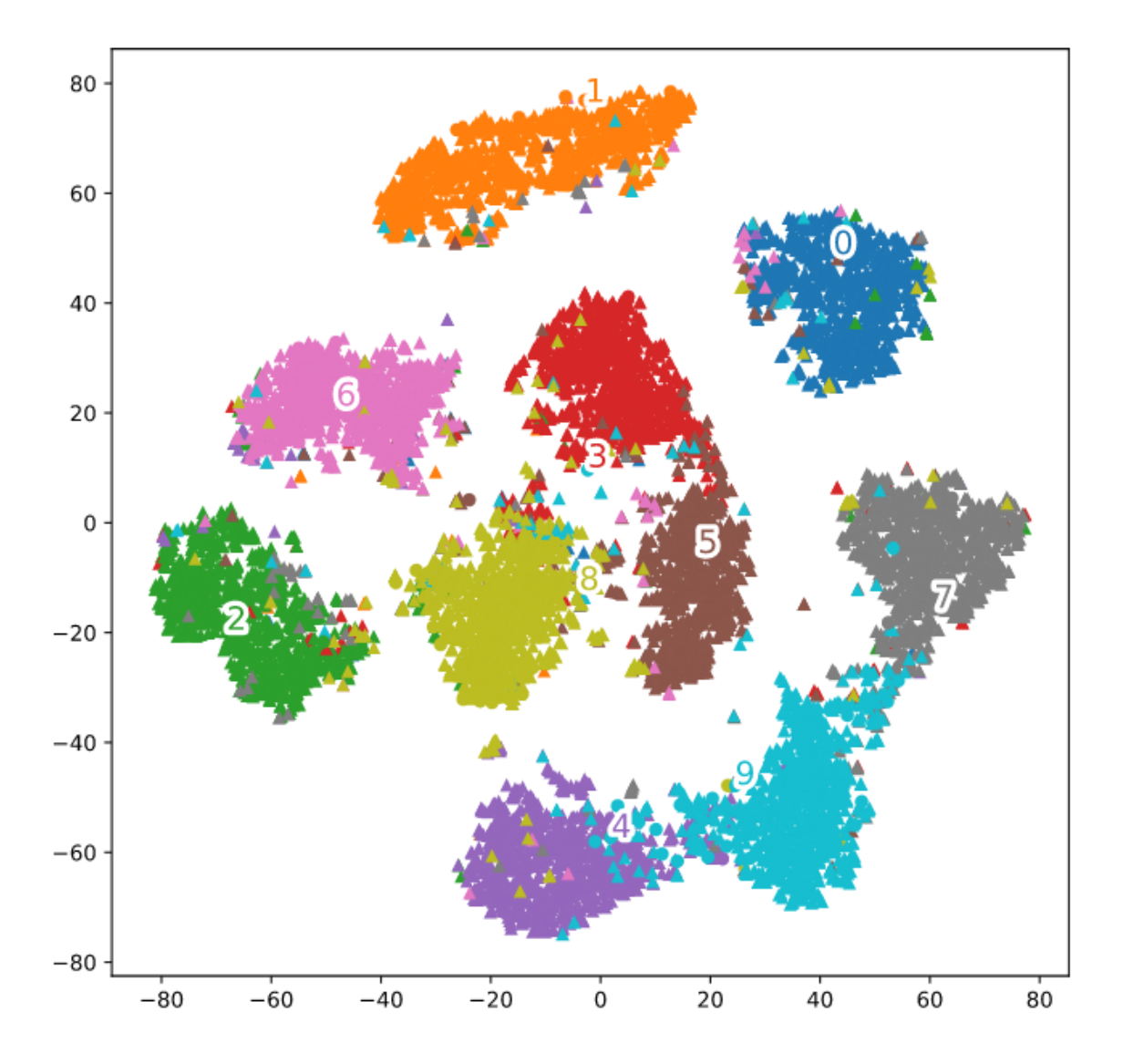} 
        \label{fig:sub1}
    \end{subfigure}
    \hspace{1mm} 
    \begin{subfigure}{0.45\textwidth}
        \centering
        \includegraphics[width=0.85\textwidth, height=6.cm]{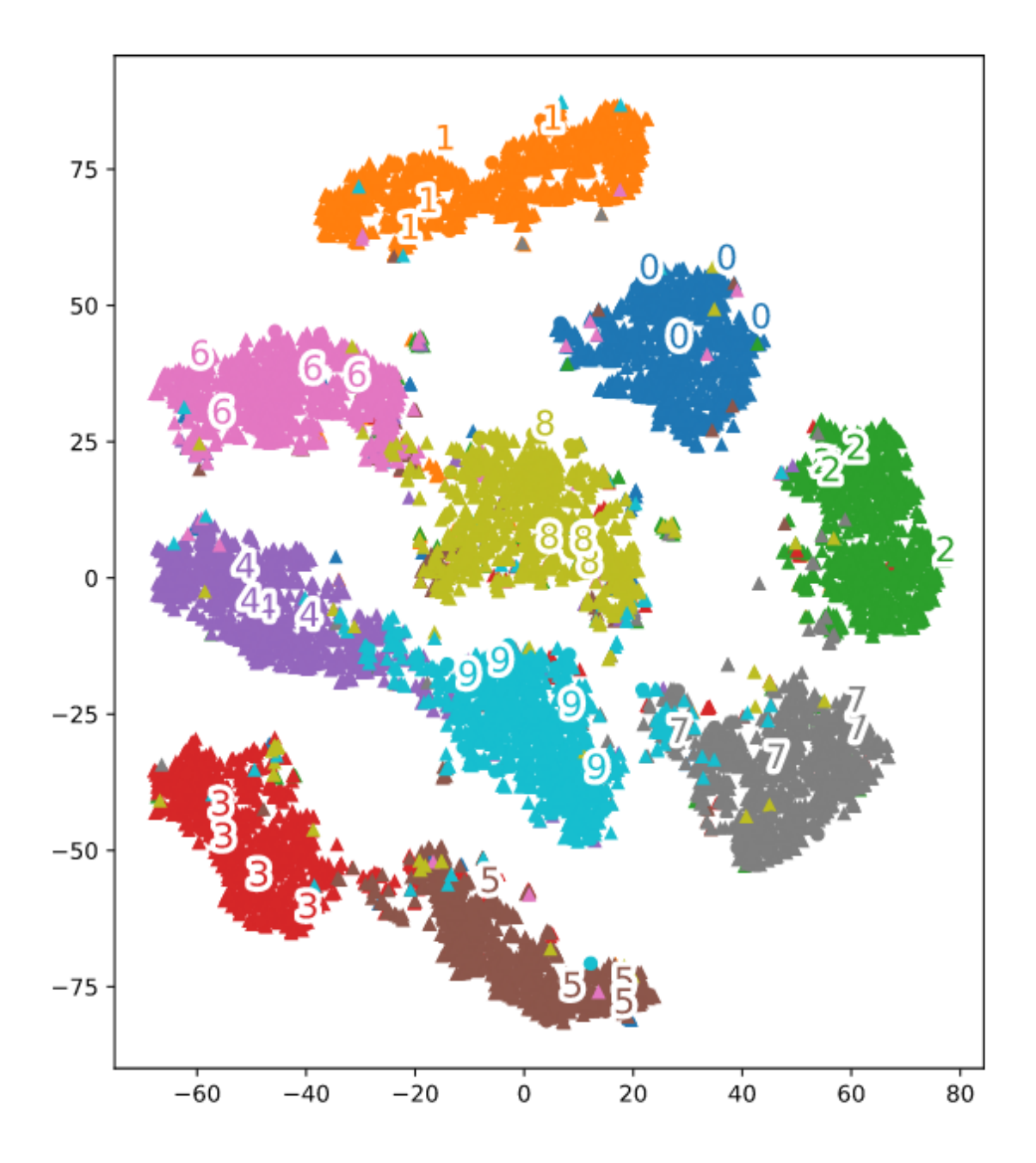} 
        \label{fig:sub2}
    \end{subfigure}
    \vspace{-5mm}
    
    \caption{\textbf{Motivation of our method} ({$t$-SNE visualization on MNIST}):  \textbf{Left}: the test latent representation of with \textit{one adaptive centroid }(i.e., visualized by digits) per class. \textbf{Right}: the test latent representation of our OTC with \textit{four adaptive centroids} per class. \textit{The centroids are learned from training samples}. \textbf{Motivation:} Based on insights from previous work \citep{DBLP:conf/wacv/0002ZZZLLM24} that identified a shift between the test and train representations, we found that using adaptive centroids is necessary to train the model in OCIL. However, using single-adaptive centroids for each classes like existing work \citep{pmlr-v162-guo22g, De_Lange_2021_ICCV} is not enough because the incoming data stream of each class is naturally multimodal, which will limit model performance if these centroids are used in training and testing later. This motivate us to our multiple-adaptive centroids as a more advanced solution.}
    \label{fig:ot-gmm}
\end{figure*}

\begin{figure*}[t]
    \centering
     \includegraphics[width=0.9\textwidth, trim=.cm 0cm 0.cm 0cm,clip]{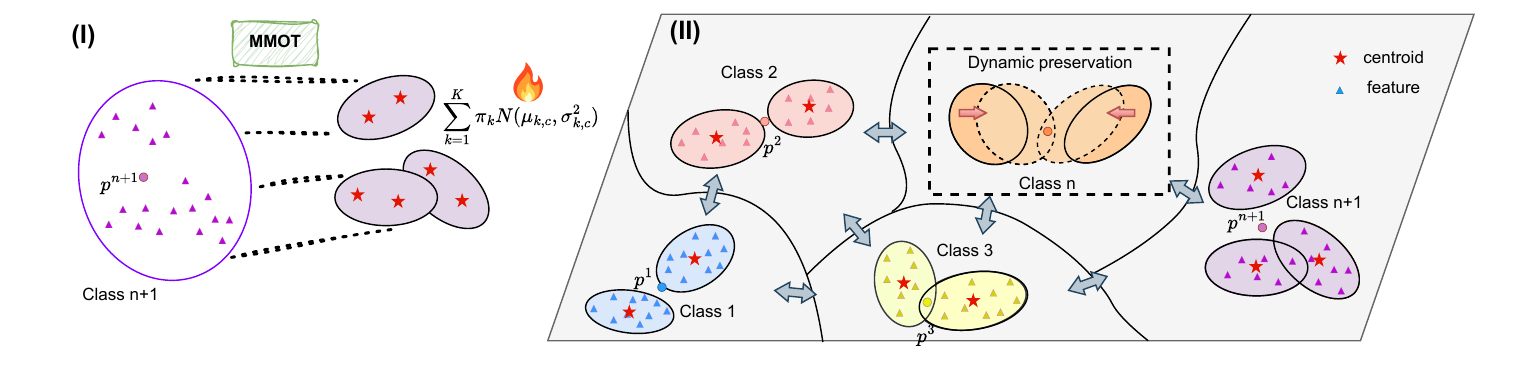}
     \vspace{-5mm}
    \caption{\textbf{Overview of our framework OTC}: \textbf{\textit{(I)}} Firstly, we perform our MMOT to incrementally characterize each class with a mixture model and multiple centroids, over time. Specifically, our MMOT will map data representations in latent space with the corresponding GMM, which we need to learn for representing the data, in an online manner. Building on this, \textbf{\textit{(II)}} we apply our Dynamic preservation with memory-buffer selection strategy to strengthen the representation learning of the Online Class Incremental learning model. Representations belonging to the same class will be pulled closer together, and conversely representations of different classes will be pushed further apart.}
    \label{fig:overview}
    \vspace{-4mm}
\end{figure*}

Artificial neural networks have sparked a revolution in addressing real-world challenges, particularly in computer vision \citep{DBLP:conf/cvpr/RedmonDGF16, DBLP:conf/nips/GoodfellowPMXWOCB14, le2025one, dao2024high, dao2025improved, dao2025self, dao2025discrete, phung2024dimsum}. The advancement of these networks has facilitated the widespread implementation of intelligent systems across various domains, introducing new and complex issues. Real-world challenges—such as autonomous vehicles \citep{MARTINEZDIAZ2018275, DBLP:journals/tits/XiaoCGUL22}, sensory robot data \citep{inproceedings, 1554340}, video streaming \citep{article, 7778898}, recommendation \citep{TRAN202289, pmlr-v101-phan-tuan19a} — require continuous interaction, learning, and adaptation from intelligent systems to effectively navigate dynamic environments. In response to this demand, Continual Learning has emerged as a promising research direction. 

\textit{Continual Learning (CL)} focuses on adapting models to changing data distributions or new tasks overtime while preventing \textit{catastrophic forgetting} \citep{robins1995catastrophic, tran-etal-2024-preserving, le2024mixture}. In this research domain, the most challenging and realistic scenario is \textit{Online Class Incremental Learning (OCIL)}, where data distribution changes dynamically, the model can only make single-iteration updates when each small batch arrives, and task IDs are unavailable during inference. Dealing with this setting, existing methods \citep{De_Lange_2021_ICCV, DBLP:conf/cvpr/GuoL023, hacohen2025predicting} often use a single classification head or centroid in latent space for each class, which may fail to capture the complexity of multimodal data, as \textit{a single class can consist of many clusters} \citep{books/mit/06/CSZ2006}. Other methods use Gaussian Mixture Models (GMMs) \citep{reynolds2009gaussian} to represent each class \citep{wang2023hide, DBLP:conf/wacv/0002ZZZLLM24}, but their means and variances are kept fixed and not updated. This makes the class representation become inaccurate, as the backbone always needs to adapt to the incoming data, leading to feature shift in latent space over time \citep{DBLP:conf/wacv/0002ZZZLLM24}. See Figure \ref{fig:ot-gmm}.

{To address these drawbacks, we propose a novel online Mixture Model based on Optimal Transport theory (MMOT) to dynamically characterize incoming streaming data. This is achieved by leveraging the rich theoretical body of Optimal Transport (OT) or Wasserstein distance \citep{villani2009optimal, santambrogio2015optimal}, along with GMM  \citep{reynolds2009gaussian}, in a specialized manner for the OCIL environment. Specifically, by employing the entropic dual form of OT \citep{genevay2016stochastic} and Gumbel softmax distribution \citep{jang2016categorical}, we reach an appealing formulation in the expectation form for the WS distance of interest, making it ready for this challenging setting where the update is performed on incoming data mini-batches. Consequently, we can incrementally capture multiple centroids of each class. As shown in Figure \ref{fig:overview}, by utilizing the multiple-centroids obtained from MMOT, our proposed Dynamic Preservation will enhance the model's class discrimination ability. 

\textbf{Contribution.} We introduce a method named \textbf{\textit{"An \underline{O}ptimal \underline{T}ransport-driven Approach for \underline{C}ultivating Latent Space in Online
Incremental Learning (OTC)"}}. Our main contributions are summarized as follows:
\begin{itemize}
    \item This work leverages Optimal Transport theory with a Gaussian Mixture model to propose a novel MMOT formulation for tackling the complexity of incoming data streams in the Online Class Incremental Learning (OCIL) scenario. Building on this foundation, our training and testing techniques not only dynamically enhance class discrimination to mitigate \textit{catastrophic forgetting}, but also narrow the gap between training and testing latent representations, ultimately improving model performance
    Building on this, our training and testing techniques not only dynamically strengthen the class discrimination ability to relieve \textit{catastrophic forgetting}, but also reduce the gap between train and test latent representations, eventually improving model performance.

    \item To our knowledge, this is the first work to explore the practical application of Optimal transport-for Mixture Model in OCIL environment. Notably, in our framework, OT not only facilitates GMM inversion but also replaces the traditional Expectation-Maximization (EM) algorithm with gradient descent. This innovation leads to significant cost savings and represents a breakthrough in handling environments where data is constantly changing.
    \item Throughout experiments on commonly used benchmark datasets, we demonstrate that our method is not only a practical solution to the multimodality of streaming data in OCIL but also achieves strong control of forgetting of previously observed tasks.
    
\end{itemize}
\section{Related work}
\label{sec:related_work}

\paragraph{Continual Learning (CL)} Generally, previous works attempt to tackle the problem of \textit{catastrophic forgetting} in CL in three main ways: \textbf{(I) Regularization-based approaches} encourage important parameters of old tasks to lie in their close vicinity \citep{kirkpatrick2017overcoming, DBLP:conf/iclr/LooST21, dao2024lifelong} by penalizing their changes.  \textbf{(II) Architecture-based  methods}  dynamically allocate  a separate subnetwork for each task \citep{DBLP:journals/corr/RusuRDSKKPH16, pmlr-v162-kang22b, tran2025boosting} to  maintain the knowledge of old tasks.
And \textbf{(III) Memory-based approaches} utilize episodic memory to store past data \citep{chaudhry2018efficient, hacohen2025predicting, thanh2025few, anh2025mutual} or employ deep generative models \citep{shin2017continual, egorov2021boovae} to produce pseudo samples from the previous history.  Among these three main lines of work, methods for OCIL, including ours, fall into the memory-based approach category, which utilizes the advantages of buffer memory to preserve the discriminative characteristics of data so far, effectively reducing \textit{catastrophic forgetting} in this challenging scenario. 

\textbf{Optimal transport for Gaussian mixture model.} \citep{altschuler2021averaging, nguyen2024barycenter} introduced closed forms for Optimal Transport (OT) and Unbalanced Optimal Transport (UOT) distance between two Gaussians. And the first OT formulation between two Gaussian Mixture Models (GMMs) was introduced by \citep{chen2018optimal}, who approached the problem by discretizing the densities and solving the resulting discrete OT problem. This framework was extended by \citep{delon2020wasserstein}, who addressed the issue that Wasserstein geodesics between GMMs generally do not remain within the space of GMMs. They proposed a variant of the Wasserstein distance that restricts the transport plans to remain within the GMM class. For high-dimensional distributions, \citep{kolouri2018sliced} developed a projected version of OT, known as Sliced Wasserstein Distance, tailored for GMMs. Alternatively, \citep{oh2023optimal} introduced a kernel-based OT-GMM method. Building on these theoretical foundations, applications in deep learning have been explored in domain adaptation by \citep{montesuma2024optimal, montesuma2024lighter} and in generative modeling by \citep{gaujac2021improving}. However, these existing work focuses on the application of \textit{forward} problem—computing distances between given GMMs—while the application of \textit{inverse} problem, namely learning GMM parameters via OT, remains largely unexplored. Our work is among the first to address this promising direction by leveraging OT for GMM parameter learning, and especially for the setting of Online Class Incremental Learning, which the related concept has not considered.

\paragraph{On dealing with Online Class Incremental Learning.}

\vspace{-4mm}
Recent work typically focuses on: either (i) how to choose meaningful, diverse observed samples to store and relay \citep{DBLP:conf/icml/ChrysakisM20, hacohen2025predicting}
(ii) or how to learn representation effectively \citep{achille2018life, rao2019continual, Gu_2022_CVPR, DBLP:conf/cvpr/YanWMZ24}, mostly inspired by contrastive learning or generative play, to mitigate forgetting.
{ Looking closer to the literature, \textit{from the perspective of OT theory for CL}, several work has been proposed, including a strategy for training VAE to generate memory buffers \citep{DBLP:conf/eccv/0004B22} or for knowledge distillation between previous models to the current learning one \citep{ye2020learning, 10.1145/3474085.3475306,10070745, dao2024lifelong}.
Different from that, \textit{we introduce a novel approach to learn and adapt multiple centroids that characterize class data in latent space}. \textit{From the view of Gaussian mixture models}, recent work in CL consider representing class data by multiple centroids obtain from this framework \citep{wang2023hide, DBLP:conf/wacv/0002ZZZLLM24}. However, they only apply the traditional version of GMM with Expectation-Minimization (EM) algorithm, and the centroids after learning are kept fixed, which then poses limitations as the model's latent space always has feature shifts when adapting to new data \citep{DBLP:conf/wacv/0002ZZZLLM24}. Notably, \textit{our MMOT framework not only addresses this drawback by flexibly updating these centroids via a simpler gradient descent algorithm, instead of EM, which always requires multiple iterations for each learning step}.} Related to using mixture model, another work \cite{achille2018life} proposed Cross-Domain Latent Homologies to characterizes data of all tasks so far with shared information. However, this can result in information loss during both data aggregation and reconstruction process, especially for highly complex datasets in current benchmark, which eventually hinders model performance. On the contrary, we are more advanced because of our adaptive version of GMM for each class, thereby limiting information loss. In addition, these centroids also participate in training backbone, helping to improve model performance.
\section{Background}
\label{sec:background}

\subsection{Optimal transport and Wasserstein distance}
Consider two distributions $\mathbb{P}$ and $\mathbb{Q}$ which operate on the domain $\Omega\subseteq\mathbb{R}^{d}$, let $d\left(\bx,\by\right)$
be a non-negative and continuous cost function or metric. Wasserstein (WS)
distance \citep{santambrogio2015optimal,villani2009optimal} between
$\mathbb{P}$ and $\mathbb{Q}$ w.r.t the metric $d$ is defined as 
\begin{equation}
\mathcal{W}_{d}\left(\mathbb{Q},\mathbb{P}\right):=\min_{\gamma\in\Gamma\left(\mathbb{Q},\mathbb{P}\right)}\mathbb{E}_{\left(\bx,\by\right)\sim\gamma}\left[d\left(\bx,\by\right)\right],\label{eq:primal_form}
\end{equation}
{where $\Gamma\left(\mathbb{Q},\mathbb{P}\right)$ is the set of couplings that admit $\mathbb{Q},\mathbb{P}$ as its marginals. }

\subsection{Entropic dual-form for OT and WS distance}\label{sec:entropic_WS}
To enable the application of OT in machine learning
and deep learning, \citep{genevay2016stochastic} developed an entropic regularized
dual form. First, they proposed to add an entropic regularization
term to primal form (\ref{eq:primal_form}) as follows:
\begin{align} \label{eq:entropic_primal}
\mathcal{W}_{d}^{\varepsilon}\left(\mathbb{Q},\mathbb{P}\right) & :=\nonumber \\
\min_{\gamma\in\Gamma\left(\mathbb{Q},\mathbb{P}\right)} & \left\{ \mathbb{E}_{\left(\bx,\by\right)\sim\gamma}\left[d\left(\bx,\by\right)\right]+\varepsilon D_{KL}\left(\gamma\Vert\mathbb{Q}\otimes\mathbb{P}\right)\right\}, 
\end{align}
where $\varepsilon$ is the regularization rate, $D_{KL}\left(\cdot\Vert\cdot\right)$
is the Kullback-Leibler (KL) divergence, and $\mathbb{Q}\otimes\mathbb{P}$
represents the specific coupling in which $\mathbb{Q}$ and $\mathbb{P}$ are independent. Using Fenchel-Rockafellar theorem \citep{tyrrell1970convex}, they obtained the following \emph{entropic regularized dual form }of (\ref{eq:entropic_primal}) as follows:
\begin{align}
\mathcal{W}_{d}^{\varepsilon}\left(\mathbb{Q},\mathbb{P}\right) & =\max_{\phi}\left\{ \int\tilde{\phi}\left(\bx\right)\mathrm{d}\mathbb{Q}\left(\bx\right)+\int\phi\left(\by\right)\mathrm{d}\mathbb{P}\left(\by\right)\right\} \nonumber \\
 & =\max_{\phi}\left\{ \mathbb{E}_{\mathbb{Q}}\left[\tilde{\phi}\left(\bx\right)\right]+\mathbb{E}_{\mathbb{P}}\left[\phi\left(\by\right)\right]\right\} ,\label{eq:entropic_dual}
\end{align}
where $\tilde{\phi}\left(\bx\right):=-\varepsilon\log\left(\mathbb{E}_{\mathbb{P}}\left[\exp\left\{ \frac{-d\left(\bx,\by\right)+\phi\left(\by\right)}{\varepsilon}\right\} \right]\right)$  {with $\phi: \Omega \to \mathbb{R}$}. \\
Please refer to \textbf{\textit{Supplementary 7}} for the background of Gaussian mixture model.
\section{Proposed Method}
\label{sec:framework}

In this section, we present the details of our proposed method. We start with the \textbf{\textit{A. General framework and motivations}}, followed by the technical details of \textbf{\textit{B. Our training strategy}}, including \textbf{\textit{MMOT}} framework, the \textbf{\textit{Dynamic preservation}} and the corresponding \textit{\textbf{memory-buffer}} techniques, which help in manipulating the latent space to maintain model performance on all data so far. Finally, we present \textbf{\textit{C. Our testing strategy}}, which helps enhance model predictive performance.

\subsection{General Framework and  Motivations}

In Online Class Incremental learning (OCIL), at each time step, our system receives a batch of new data samples $X = [X^c]_{c\in\mathcal{C}{new}}$, where $\mathcal{C}_{new}$ represents classes of new data and $X^c$ is the batch data for class $c$. To mitigate catastrophic forgetting, we maintain a memory buffer $\mathcal{M}$ of the old classes encountered so far. Therefore, during the training of new data, we randomly retrieve a batch of old data $\bar{X} = [\bar{X}^c]_{c \in \mathcal{C}{old}}$ from $\mathcal{M}$, where $\mathcal{C}_{old}$ represents the observed classes. During training, we feed $X$ and $\bar{X}$ to the feature extractor $f_{\boldsymbol{\theta}}$ to obtain the batches of latent representations $Z =f_{\boldsymbol{\theta}}(X)$ and $\bar{Z} =f_{\boldsymbol{\theta}}(\bar{X})$ for new and old data, respectively.

Traditionally, most existing works \citep{De_Lange_2021_ICCV, DBLP:conf/cvpr/GuoL023, hacohen2025predicting} have used a single prototype to represent each class, applying an objective function to pull feature vectors of the same class toward the prototype and push them away from other class prototypes. This strategy effectively reduces intra-class separation and increases inter-class separation, achieving good performance. However, using a single prototype may not capture the complexity of incoming data, as practical data often exhibits multimodality, where a class may consist of many clusters \citep{books/mit/06/CSZ2006}. Thus, it may not adequately generalize a class, limiting model performance, as shown in Figure \ref{fig:ot-gmm}. Recent works \citep{wang2023hide, DBLP:conf/wacv/0002ZZZLLM24} use multiple centroids via Gaussian Mixture Models (GMMs) to characterize latent space, but GMMs using the EM algorithm require many expensive iterations. Moreover, these methods simply
calculate these centroids once and keep them fixed, leading to diminishing representativeness as features shift when models adapt to new data \citep{DBLP:conf/wacv/0002ZZZLLM24}. \textit{These motivate our training strategy, which can effectively learn and adapt multiple centroids for each data class,} (see Figure \ref{fig:overview}) as follow:
\begin{itemize}
    \item For the batches of $X$ and $\bar{X}$, we first perform some initial training steps using Cross Entropy (CE) Loss, making samples in the same classes closer and samples from different classes more separate. 
    \item Subsequently, given the initial separation of the old and new classes, we perform our MMOT framework, which \textit{incrementally estimates the distribution for each class over time,} to tackle the complexity of incoming data streams. {Notably, our MMOT update centroids with several cheap gradient descent steps}.
    \item Based on that, we introduce a complementary component, named the \textit{Dynamic preservation}. This component leverages information from the distributions learned from MMOT to enhance the representation learning efficiency of models.
\end{itemize}
Algorithm~\ref{alg:main_fw} summarizes the main steps of our method, as also illustrated in Figure~\ref{fig:overview}. In what follows, we present and discuss the technicality of our MMOT and Dynamic preservation technique. 

\begin{algorithm}[tp]
\caption{Our training strategy (OTC)}
\textbf{Input:}{ The batches $X = [X^c]_{c\in \mathcal{C}_{new}}$ and $\bar{X} = [\bar{X}^c]_{c \in \mathcal{C}_{old}}$}. \\
\textbf{Output:}{Feature extractor $f_{\boldsymbol{\theta}}$, the centroids/ covariance matrices $[\boldsymbol{\mu}^c_k, \Sigma^c_k]_{k=1}^K$ for each class $c$}
\begin{algorithmic}[1]
\FOR{each batch $(X, \bar{X})$}
    \STATE Step 0. Perform initial training with CE loss.
    \STATE Step 1. Perform MMOT (Algorithm \ref{alg:ot_gmm})
    \STATE Step 2. Perform dynamic preservation
    \STATE Step 3. Update the replay memory $\mathcal{M}$
\ENDFOR
\end{algorithmic}\label{alg:main_fw}
\end{algorithm}

\subsection{Our training strategy}

\subsubsection{Multimodality with Optimal transport (MMOT)}

\paragraph{a. The Derivation of MMOT:}

This is the key module of our framework. Given a class $c$, we aim to exploit the mature theoretical body of OT to develop the online algorithm, where the centroids and covariance matrices of the corresponding mixture model are incrementally updated according to incoming data streams.  

Let the latent representations or feature vectors of the class $c$ be $D_c=\left\{ \bz_1^c,...,\bz^c_{N_c}\right\}$, wherein each $\bz_i^c = f_{\boldsymbol{\theta}}(\bx_i^c)$ is the representation of data sample $\bx_i^c$ in the data stream. We denote $\mathbb{P}_{c}$
 as the empirical data distribution of latent representations of class $c$. We need to learn a Gaussian mixture model (GMM) that approximates the data distribution $\mathbb{P}_{c}$. Consider the following GMM: 
\begin{equation} \label{eq:gmm}
\mathbb{Q}_c:=\sum_{k=1}^{K}\pi_{k,c}\mathcal{N}\left(\bmu_{k,c},\diag\left(\vectorize{\sigma}_{k,c}^{2}\right)\right),
\end{equation}
where $\pi_{k,c}$ is mixing proportion, $\bmu_{k,c} \text{ and } \diag\left(\vectorize{\sigma}_{k,c}^{2}\right)$
are the mean vector and covariance matrix of the $k$-th Gaussian. To learn this GMM, we propose minimizing WS distance between $\mathbb{P}_{c}$
and $\mathbb{Q}_{c}$ as follows:
\begin{equation}
\min_{\bpi^c,\bmu^c,\Sigma^c}\mathcal{W}_{d}\left(\mathbb{P}_{c},\sum_{k=1}^{K}\pi_{k,c}\mathcal{N}\left(\bmu_{k,c},\diag\left(\vectorize{\sigma}_{k,c}^{2}\right)\right)\right),
\label{obj}
\end{equation}
where $\bpi^c=\left[\pi_{k,c}\right]_{k=1}^{K}$, $\bmu^c=\left[\bmu_{k,c}\right]_{k=1}^{K}$,
$\Sigma^c=\left[\diag\left(\vectorize{\sigma}_{k,c}\right)\right]_{k=1}^{K}$, and $d$ is a distance on the latent space. {In this line of thought, the existence of an optimal solution to the optimization problem (\ref{obj}) has also been established by \citep{lasserre2024gaussian}}. \\
Now in order to handle the above WS distance, we need to sample $\tilde{\bz}^c \sim \mathbb{Q}_c=\sum_{k=1}^{K}\pi_{k,c}\mathcal{N}\left(\bmu_{k,c},\diag\left(\vectorize{\sigma}_{k,c}^{2}\right)\right)$,
using the re-parameterization trick, {which makes sampling differentiable for gradient-based optimization} as follows:
\begin{equation}
\tilde{\bz}_{k}^c=\bmu_{k,c}+\bepsilon_{k}\diag\left(\vectorize{\sigma}_{k,c}\right),
\end{equation}
where the source of randomness $\bepsilon_{k}\sim\mathcal{N}\left(\mathbf{0},\mathbb{I}\right)$. We then sample the one-hot vector $\bl= [\l_{k}]_{k=1}^K\sim\text{Cat}\left(\bpi^c\right)$
and compute 
\begin{equation}
\tilde{\bz}^c=\sum_{k=1}^{K}\l_{k}\tilde{\bz}_k^c=\sum_{k=1}^{K}\l_{k}\left(\bmu_{k,c}+\bepsilon_{k}\diag\left(\vectorize{\sigma}_{k,c}\right)\right).
\end{equation}
However, to do a continuous relaxation \citep{jang2016categorical, maddison2016concrete} of $\bl$ for enabling learning
$\bpi^c$ {through} gradient descent updating, we {use Gumbel-Softmax trick for differentiable sampling from categorical component} as follows: 
\begin{align*}
\l_{k} & =\frac{\exp\left\{ \left(\log\pi_{k,c}+G_{k}\right)/\tau\right\} }{\sum_{j=1}^{K}\exp\left\{ \left(\log\pi_{j,c}+G_{j}\right)/\tau\right\} }, k=1,...,K\\
\text{Thus, } \tilde{\bz}^c &=\sum_{k=1}^{K}y_{k}\tilde{\bz}_{k}^c=\sum_{k=1}^{K}y_{k}\left(\bmu_{k,c}+\bepsilon_{k}\diag\left(\vectorize{\sigma}_{k,c}\right)\right),
\end{align*}
where $\tau>0$ is a temperature parameter and random noises $G_{k}$
are i.i.d. sampled from Gumbel distribution (i.e., $G_{k}=-\log\left(-\log u_{k}\right)$
for $u_{k}\sim\text{Uniform}(0,1)$). Finally, to capture the data distribution of each class $c$ (i.e., solving (\ref{obj})), we use the $\varepsilon$-entropic dual-form \citep{genevay2016stochastic} as 
\begin{align}
 & \mathcal{W}^{\varepsilon}_{d}\left(\mathbb{P}_{c},\sum_{k=1}^{K}\pi_{k,c}\mathcal{N}\left(\bmu_{k,c},\diag\left(\vectorize{\sigma}_{k,c}^{2}\right)\right)\right)\nonumber \\
 & =\max_{\phi}\left\{ \mathbb{E}_{\mathbb{P}_{c}}\left[\phi\left(\bz^c\right)\right]+\mathbb{E}_{\mathbb{Q}_{c}}\left[\tilde{\phi}\left(\tilde{\bz}^c\right)\right]\right\} ,\label{eq:dual_OT_GMM}
\end{align}
where $\varepsilon>0$ is a small number, $\phi$ is the Kantorovich network, $\tilde{\bz}^c=\sum_{k=1}^{K}y_{k}\left(\bmu_{k,c}+\bepsilon_{k}\diag\left(\vectorize{\sigma}_{k,c}\right)\right)$, and 
\[
\tilde{\phi}\left(\tilde{\bz}^c\right)=-\varepsilon\log\left(\mathbb{E}_{\mathbb{P}_{c}}\left[\exp\left\{ \frac{-d\left(\bz^c,\tilde{\bz}^c\right)+\phi\left(\bz^c\right)}{\varepsilon}\right\} \right]\right).
\]

\paragraph{b. MMOT in {Online Class Incremental learning}:}

We now present how to perform our MMOT in the online continual learning scenario when we need to gradually update the set of centroids and covariance matrices for each class $c$ based on the batch $X^c$ or $\bar{X}^c$. Looking into Eq. (\ref{eq:dual_OT_GMM}), this objective function is in the form of expectation, hence perfectly fitting for online learning. Specifically, we use the current batch $X^c$ or $\bar{X}^c$ for each class $c$ to solve:
\begin{align}
\min_{\bpi^c,\bmu^c,\Sigma^c}\max_{\phi}\left\{ \mathbb{E}_{X^{c}\,or\,\bar{X}^{c}}\left[\phi\left(f_{\boldsymbol{\theta}}(\bx^c)\right)\right]+\mathbb{E}_{\mathbb{Q}_{c}}\left[\tilde{\phi}\left(\tilde{\bz}^c\right)\right]\right\} .\label{eq:dual_OT_GMM-batch}
\end{align}
To solve the optimization problem (\ref{eq:dual_OT_GMM-batch}) for each class $c$, we update the Kantorovich network $\phi$ few times and then gradually update the mixing proportions $\bpi^c$, the set of centroids $\bmu^c$, and the set of covariance matrices $\Sigma^c$ for class $c$ via the gradient descent algorithm.  The key steps of MMOT is summarized in Algorithm~\ref{alg:ot_gmm}. Note that although the GMMs for different classes are learned independently, each mixture is updated only with its own class data streamed through the buffer. Thus, even though the OT optimization is unsupervised within a class, the process remains class-conditional globally, avoiding uncontrolled mixing across classes. Eventually, we obtain the centroids $[\bpi^c]_{c \in \mathcal{C}}$ and covariance matrices $[\Sigma^c]_{c \in \mathcal{C}}$, which then are used for the other following strategies, including \textit{\textbf{Dynamic preservation}}, 
selecting diverge \textbf{\textit{memory-bufer}} and finally effectively making inference in the \textit{\textbf{testing phase}}. 

\begin{algorithm}[ht] 
\caption{MMOT in OCIL scenario}
\textbf{Input:}{The batches $X = [X^c]_{c\in \mathcal{C}_{new}}$ and $\bar{X} = [\bar{X}^c]_{c \in \mathcal{C}_{old}}$} \\
\textbf{Output:}{$[\bpi^c, \bmu^c, \Sigma^c]_{c \in \mathcal{C}}$}
\begin{algorithmic}[1]
    \FOR{ each $c \in \mathcal{C}$}
      \STATE Update $\phi$ according to (\ref{eq:dual_OT_GMM-batch}).
      \STATE Update $[\bpi^c, \bmu^c, \Sigma^c]$ according to (\ref{eq:dual_OT_GMM-batch}).    
    \ENDFOR
\end{algorithmic}
\label{alg:ot_gmm}
\end{algorithm}
\vspace{-4mm}
\paragraph{c. Discussion:} \textbf{\textit{Regarding the advantages of our method:}} Compared to existing work, MMOT enables us to use multiple adaptive centroids to characterize a class \textit{in an online learning manner}. Figure~\ref{fig:ot-gmm} visually demonstrates our advantages. Specifically, we use t-SNE to visualize the test latent representations of the current approach, using adaptive single-centroid, and our MMOT when utilizing \textit{adaptive multiple-centroids} per class. We observe that there exists a shift between test and train representations. Hence, the centroids tailored to the train set might mismatch the test set. However, it can be seen that using multiple centroids per class, as in our MMOT, can mitigate this mismatch. \\
\textbf{\textit{Regarding why we use Optimal transport (OT) instead of Kullback–Leibler (KL) divergence?}} {One of the central ideas in our approach is to learn the parameters of a Gaussian Mixture Model (GMM) by minimizing a distance between distributions. We opt for the Wasserstein distance \citep{arjovsky2017wasserstein} from OT, rather than KL, a more common alternative. This is because (i) \citep{kolouri2018sliced} showed that minimizing the KL divergence asymptotically corresponds to maximizing the log-likelihood via the Expectation-Maximization (EM) algorithm when the number of samples grows to infinity.  Moreover, EM, which requires multiple iterations, is often costly for an online algorithm (more discussion in Supplementary). Furthermore, the Wasserstein distance \citep{arjovsky2017wasserstein} offers other compelling advantages: (ii) it is a proper and continuous metric that is differentiable everywhere; (iii) it maintains numerical stability even when the distributions have minimal or disjoint support; and (iv) unlike divergences such as KL that disregard the underlying geometry of the data, the Wasserstein distance respects the structure and spatial relationships of the distributions, leading to more faithful approximations in GMM learning.
}

\begin{table*}[!ht]
    \centering
    
    \caption{\textbf{Average Accuracy (higher is better)}, M denotes the memory buffer size. All numbers are the average of 5 runs. The data in the table represents Average Accuracy ± standard deviation.}
    \vspace{-3mm}
    \renewcommand\arraystretch{1.12}
    \small{
    \resizebox{0.65\textwidth}{!}{\begin{tabular}{l rrr|rrr|rrr}
        \toprule 
        Method & M = 0.2k & M = 0.5k & M = 1k & M = 1k & M = 2k & M = 5k & M = 2k & M = 5k & M = 10k \\
        \midrule
         \midrule
        {ER} & 51.2±0.9 & 56.7±1.9 & 62.3±4.1 & 24.5±0.7 & 31.9±1.5 & 39.4±1.8 & 10.8±0.8 & 19.2±1.4 & 24.7±2.5  \\
        {ASER} & 27.8±1.0 & 36.2±1.2 & 44.7±1.2 & 16.4±0.3 & 12.2±1.9 & 27.1±0.3 & 5.3±0.3 & 8.2±0.2 & 10.3±0.4 \\
        {CoPE} & 37.3±2.2 & 42.9±3.5 & 43.8±3.5 & 14.6±1.3 & 16.8±0.9 & 23.6±0.9 & 2.3±0.4 & 2.5±0.3 & 2.7±0.6 \\

        {OCM} & 61.0±1.6 & 66.2±1.8 & 73.2±1.1 & 28.0±0.7 & 35.7±1.4 & 42.2±1.1 & 18.4±1.0 & 26.7±1.0 & 31.9±1.2 \\

        {GSA} & 58.0±0.4 & 64.6±0.2 & 69.1±0.3 & 31.4±0.2 & 39.7±0.6 & 49.7±0.2 & \underline{18.5±0.4} & 26.0±0.2 & 33.2±0.4 \\

        {OnPro} & 62.7±1.7 & 70.5±1.9 & 74.7±1.5 & 30.0±0.4 & 35.9±0.6 & 41.8±1.2 & 16.3±1.4 & 21.1±2.1 & 26.4±2.2 \\

        {MOSE} & 53.3±1.2 & 61.1±1.5 & 70.7±1.2 & 35.1±0.3 & 45.1±0.3 & \underline{54.5±0.8} & 18.2±0.7 & \underline{30.9±0.6} & \underline{38.7±0.4} \\

        {SBS} & 52.2±0.8 & 58.3±1.6 & 64.8±3.2 & 26.3±0.8 & 34.2±1.4 & 42.1±1.8 & 10.8±0.9 & 23.2±1.5 & 27.3±2.7 \\

        BiC+AC & \underline{63.5±2.3} & \underline{71.2±2.5} & \underline{75.8±3.1} & \underline{36.1±0.8} & \underline{47.3±1.5} & 54.2±1.8 & 17.6±1.0 & 22.6±1.2 & 26.5±1.6  \\
         
         \midrule
         OTC (Ours) & \textbf{64.8±1.6} & \textbf{72.0±1.9} & \textbf{76.1±2.0} & \textbf{36.8±1.5} & \textbf{48.5±2.1} & \textbf{56.5±1.9} & \textbf{19.5±2.0} & \textbf{31.6±2.1} & \textbf{39.5±2.4} \\
         
         \midrule
         & \multicolumn{3}{c}{a) CIFAR-10} & \multicolumn{3}{c}{b) CIFAR-100} & \multicolumn{3}{c}{c) Tiny-Imagenet}\\
          
         
        \end{tabular}}} \\
        
    \label{tab:bal_compare_acc}
\end{table*}




\begin{table*}[!ht]
    \centering
    
    \caption{\textbf{Average Forgetting (lower is better)}, M denotes the memory buffer size. All numbers are the average of 5 runs. The data in the table represents Average Forgetting $\pm$ standard deviation.}
    \vspace{-3mm}
    \renewcommand\arraystretch{1.15}
    \small{
    \resizebox{0.65\textwidth}{!}{\begin{tabular}{l rrr|rrr|rrr}
        \toprule 
        Method & M = 0.2k & M = 0.5k & M = 1k & M = 1k & M = 2k & M = 5k & M = 2k & M = 5k & M = 10k \\
        \midrule
         \midrule
        {ER} & 40.2±3.3 & 33.2±3.5 & 20.9±6.8 & 32.7±1.8 & 22.2±2.3 & 13.3±1.9 & 58.4±1.7 & 45.7±1.6 & 40.8±2.5 \\
        {ASER} & 62.6±1.1 & 53.2±1.5 & 42.0±1.2  & 50.4±0.9 & 46.8±0.7 &  45.3±0.8 & 75.4±0.7 & 67.5±1.0 & 64.3±0.9 \\
        {CoPE} &  45.7±1.5 & 39.4±1.8 & 36.4±1.5 & 17.8±1.3 & 14.4±0.8 & 12.8±1.1 & \textbf{10.9±0.4} & \textbf{9.0±0.5} & \textbf{8.2±0.6} \\

        {OCM} & 25.3±3.5 & 13.7±3.3 & 11.6±2.6 & 15.0±1.6 & \textbf{9.2±1.8} & \textbf{3.8±1.2} & 26.1±1.6 & 19.7±1.3 & \underline{15.9±1.5} \\

        {GSA} & 23.5±0.2 & \textbf{12.6±0.4} & \underline{10.0±0.3} & 33.2±0.6 & 22.8±0.4 & 8.7±0.3 & 35.5±0.3 & 25.8±0.4 & 16.9±0.6 \\

        {OnPro} & 25.9±2.1 & 17.9±2.7 & 14.2±2.6 & 16.8±2.5 & 12.4±1.4 & 6.7±0.9 & 28.0±1.6 & 23.5±1.8 & 20.3±1.7 \\

        {MOSE} & 38.5±2.1 & 30.4±1.7 & 20.3±1.3 & 37.5±0.4 & 25.9±0.5 & 13.6±0.6 & 47.2±1.4 & 25.0±0.6 & 15.5±0.3 \\

        {SBS} & 40.1±1.1 & 34.2±1.5 & 27.7±1.4 & 46.3±0.8 & 36.4±1.2 & 25.7±1.3 & 54.2±1.3 & 33.8±1.5 & 26.8±1.7 \\

        BiC+AC & \underline{23.4±1.8} & {16.4±1.5} & {12.8±1.4} & \underline{12.1±1.5} & {10.4±1.3} & \underline{5.5±1.4 }& 27.4±1.3 & 21.6±1.2 & 20.1±1.2  \\
         
         \midrule
         OTC (Ours) & \textbf{23.2±1.1} & \underline{13.5±1.2} & \textbf{9.8±0.8} & \textbf{11.3±0.9} & \underline{10.0±0.9} & {6.3±0.8} & {23.5±1.0} & {19.8±1.1} & {16.5±1.2} \\
         
         \midrule
         & \multicolumn{3}{c}{a) CIFAR-10} & \multicolumn{3}{c}{b) CIFAR-100} & \multicolumn{3}{c}{c) Tiny-Imagenet}\\
          
         
        \end{tabular}}} \\
        
    \label{tab:bal_compare_forget}
\end{table*}

\begin{figure*}[!ht]
    \centering
    \vspace{-2mm}
    \includegraphics[width=0.79\textwidth]{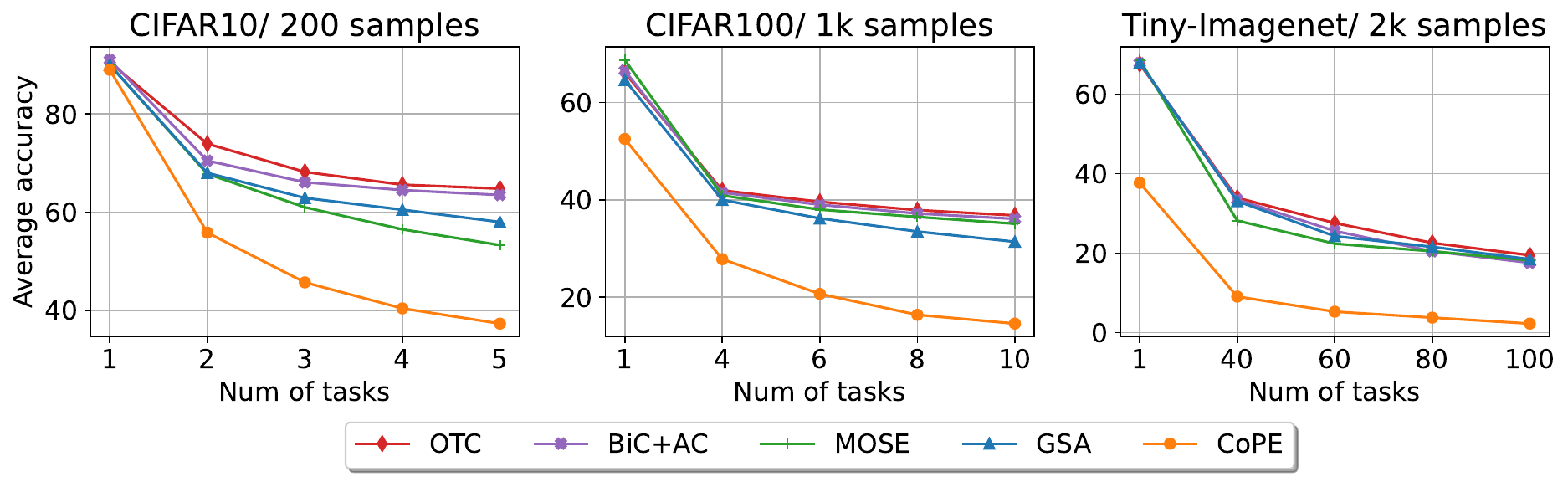}
    \vspace{-3mm}
    \caption{Average accuracy through tasks.}
    \vspace{-3mm}
    \label{fig:linesplot}
\end{figure*}

\subsubsection{Dynamic Preservation}
\label{dynamic_preserve}
Taking advantage of the learned distributions via our MMOT strategy, we introduce an objective function to improve model representation learning. In particular, for each class $c \in \mathcal{C} = \mathcal{C}_{new} \cup \mathcal{C}_{old}$, we have the corresponding combined learning batch $\mathcal{X}^c = X^{c} \cup \bar{X}^{c}$ for each learning step. In each step, instead of using just one prototype per class to define the latent space constraints, we leverage the centroids of each class from the MMOT model $G^c$ to form the objective function as follows: 
\vspace{-2mm}
\begin{equation}
\mathcal{L}_{DP}(\theta) = \mathbb{E}_{c \in \mathcal{C}}\mathbb{E}_{\bx^c\sim \mathcal{X}^c} \log \frac{g_{cen}^c}{\sum_{c'} \left( g_{cen}^{c'} + g_{fea}^{c'} \right)}
\label{obj_l}
\end{equation}
where $g_{cen}^{c'} = \sum_{k=1}^K\exp(f_{\boldsymbol{\theta}}\left(\bx^c\right) \cdot \mu_{k,c'} / \tau)$ encourage the representations of class $c$ to move as close to their respective centroids as possible, thereby making these representations of this class become closer together , and $g_{fea}^{c'} =
\exp(f_{\boldsymbol{\theta}}\left(\bx^c\right) \cdot f_{\boldsymbol{\theta}}\left(\bx^{c'}\right) / \tau)$ aim to push the centroids and features of each class $c' \neq c$ away from $c$, thereby increasing inter-class separation. According to Equation (\ref{obj_l}), using centroids instead of a single prototype is effective because the information about the classes is represented more specifically and clearly. Centroids located on the boundaries of the classes particularly help strengthen the effectiveness of learning the representation.

After performing dynamic preservation, the latent representations of the same classes become closer, while those from different classes become more separate. As a result, we increase the class discrimination ability during online incremental learning, aiding MMOT in mitigating catastrophic forgetting more efficiently. (cf. Figure~\ref{fig:overview}). In addition, we observe that although becoming more condensed, the latent representations of each class of data stream are still complex with multi-modality.
This strongly motivates us to use the centroid information obtained from MMOT for further testing and updating the memory buffer.



\subsubsection{Replay memory selection:} After processing each batch $(X, \bar{X})$, as shown in Step 3 (Line 5) in Algorithm \ref{alg:main_fw}, we select some data points to supplement the replay memory of new task as follows: 
\begin{itemize}
    \item For a centroid of a class, we choose some closest data points of this class in the current batch to add to the replay memory.
    \item If the replay memory is full, we randomly pick some data points in the current replay memory to replace the fresh-new ones.
\end{itemize}
In this way, we expect the resulting memory buffer to contain a representative samples that effectively characterize old data, thus effectively supporting the Dynamic Preservation strategy to reduce forgetting.

\subsection{Doing inference:}
Since the centroids obtained from MMOT are representative of the class data, we propose to utilize this information during inference to improve the final performance. In particular, given an unseen data point $\bx$, we compute the Mahalanobis distance $d_{MH}$ of $\bx$ to each Gaussian of $c$ and classify $\bx$ to the closest class as follows:
\begin{align*}
d\left(\mathbf{x}, c\right) &= \min_{k=1, \ldots, K} d_{MH} \left( f_{\boldsymbol{\theta}}\left(\mathbf{x}\right), \mathcal{N} \left( \boldsymbol{\mu}_{k,c}, \operatorname{diag}\left(\boldsymbol{\sigma}_{k,c}^{2} \right)\right) \right) \\
\hat{y} &=\text{argmin}_{c\in\mathcal{C}}d\left(\bx,c\right).
\end{align*}
\section{Experiment}
\label{sec:exp}
\subsection{Experimental setup}
\paragraph{Datasets:}
We use four benchmark datasets, which are widely used in OCIL: Tiny-ImageNet, CIFAR-100, CIFAR-10, and MNIST. 
\vspace{-4mm}
\paragraph{Baselines:} To demonstrate the our effectiveness in the field of OCIL, we conduct experiments on 9 typical and state of-the-art baseline methods: \textbf{ER} \citep{DBLP:conf/nips/RolnickASLW19}, \textbf{{ASER}} \citep{shim2021online}, \textbf{{CoPE}} \citep{De_Lange_2021_ICCV}, \textbf{OCM} \citep{pmlr-v162-guo22g}, \textbf{GSA} \citep{DBLP:conf/cvpr/GuoL023}, \textbf{OnPro} \citep{onpro}, \textbf{MOSE} \citep{DBLP:conf/cvpr/YanWMZ24}, \textbf{SBS} \citep{hacohen2025predicting} and \textbf{BiC+AC} \citep{szatkowski2025improving}.
\vspace{-4mm}
\paragraph{Metrics:} We use two main metrics: Final Average Accuracy (FAA) and the Final Forgetting Measure (FFM). \\
Please refer to \textbf{\textit{Supplementary \ref{sec:imp_detail}}} for the further detailed implementation.

\begin{figure*}[!ht]
    \centering
    \includegraphics[width=0.8\textwidth]{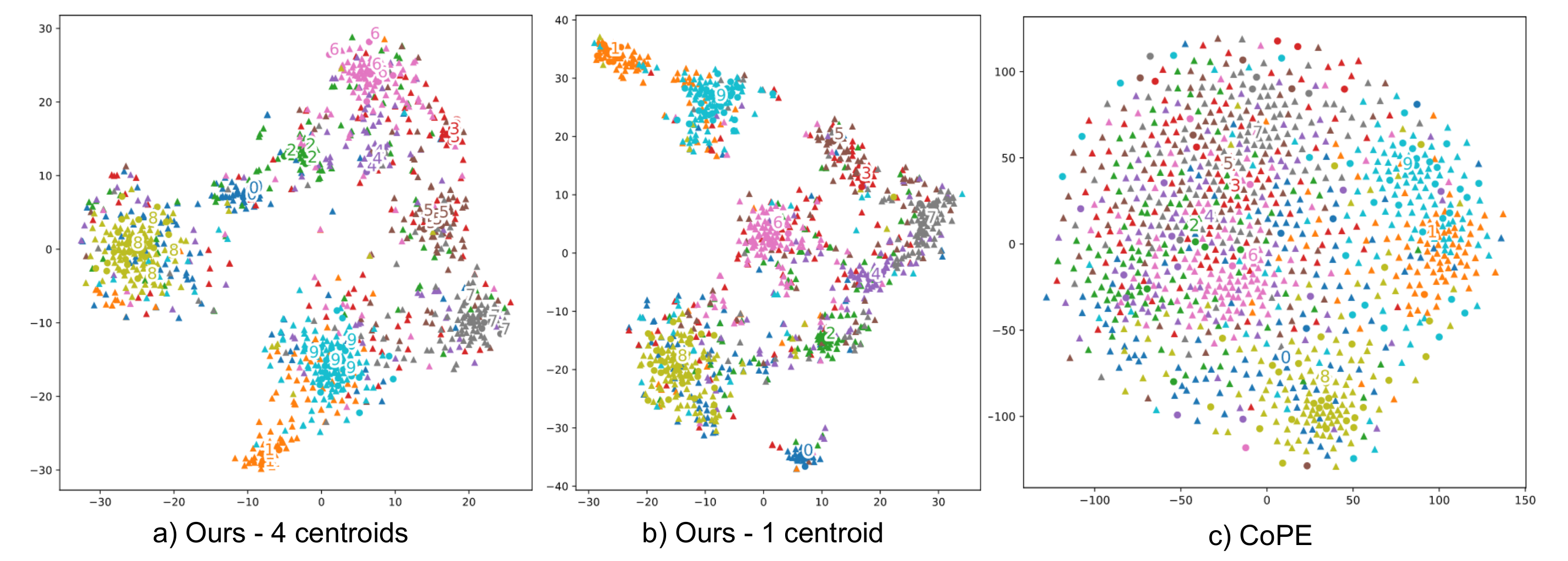}
    \vspace{-3.5mm}
    \caption{\textbf{Features on latent spaces} of our method (a and b) and CoPE (c). It can be observed that \textbf{\textit{(I)}} For our method, using 4 centroids is better than using just 1 centroid when predictions; \textbf{\textit{(II)}} OTC is always better than CoPE with 1 adaptive centroid due to the effect of our Dynamic preservation and buffer selection strategies wrt representation learning. We compare ours with CoPE to further investigate the reason for CoPE's impressive Average Forgetting reported in Table 2.}
    \label{fig:our_vs_cope}
    \vspace{-3mm}
\end{figure*}

\begin{figure}[!ht]
    \centering
    \includegraphics[width=0.48\textwidth]{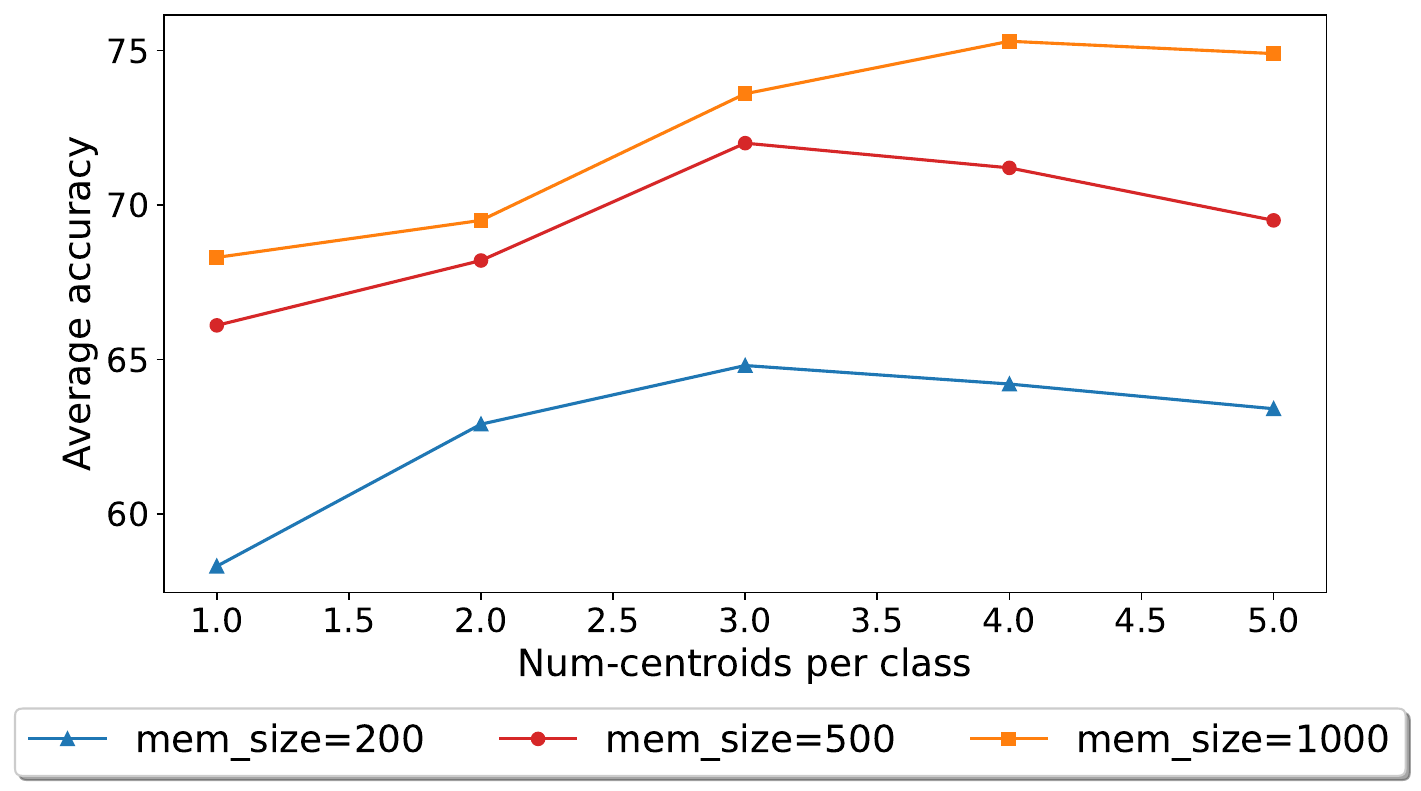}
    \vspace{-6mm}
    \caption{Accuracy by different \#centroids/class (CIFAR10).}
    \label{fig:num_cluster}
    \vspace{-5mm}
\end{figure}

\subsection{Performance comparison}


Table \ref{tab:bal_compare_acc} summarizes the final average accuracy of models on three challenging datasets with various memory sizes. In general, our OTC outperforms the baselines with a margin of up to 2\%. In addition, the performance improvement of ours is always the best when the memory bank size is the smallest for each dataset, which is critical for OCIL setting with limited resources. There is only one case when our method performs quite similarly with BiC+AC, but ours is still better than that baseline (CIFAR10 with memory size $M=1k$). However, on more challenging datasets, our method shows significant superiority, outperforming by up to 2\% and 13\% on CIFAR100 and Tiny-Imagenet, respectively. These results confirm the effectiveness of our MMOT framework for training and testing performance in the challenging environment of OCIL, especially when the model has to deal with a long sequence of tasks, as in the case of Tiny-Imagenet (i.e., 20 tasks).



In addition, Table \ref{tab:bal_compare_forget} shows the average forgetting at the end of the data stream for the methods applied to the three corresponding datasets. The results demonstrate that our OTC can effectively mitigate catastrophic forgetting. Overall, OTC consistently ranks among the top two methods with the lowest average forgetting on CIFAR10 and CIFAR100. On Tiny-Imagenet, however, our method experiences more forgetting than CoPE, with a significant gap.

To further investigate this phenomenon, we illustrate the accuracy of the models corresponding to the methods, including OTC, the best baselines (GSA, MOSE, BiC+AC), and CoPE—the baseline with the least forgetting. The results in Figure \ref{fig:linesplot} show that, in general, our method's performance curve remains at the top. The results also indicate that CoPE performs poorly on Tiny-Imagenet from the starting time, thus the difference between the initial accuracy during learning and the accuracy after learning the sequence of tasks is minimal, leading to a small final average forgetting. Additionally, Figure \ref{fig:our_vs_cope} provides a t-SNE visualization comparing the latent space of our method and CoPE on MNIST, which further confirms the differences in the representation learning quality of the two methods. Importantly, our OTC is still among the top three methods with the smallest forgetting on this challenging dataset, and our forgetting is always less than the top 2 baselines, which give the best final accuracy. These results further prove the effectiveness of our method in learning representations and avoiding forgetting compared to typical arts.

\subsection{Ablation study}


\textbf{The role of using multiple centroids in general:} For our MMOT's efficacy, we examine the influence of the number of centroids per class wrt model performance. Figure \ref{fig:num_cluster} depicts curves of average accuracies on CIFAR10 when \textit{Num-centroids} varies. In general, the performance improves when we increase the number of centroids per class from 1 to a certain threshold. In addition, the bigger the memory size is, the bigger the threshold is. Moreover, if these thresholds are exceeded, the model performance depends on the support level of replay memory: {the smaller memory size leads to the lower quality of the representation learning, and higher prediction error}. In particular, when the memory size $M=200$, if the number of centroids is bigger than 3, the model performance degrades. Whereas, with memory size $M= 1K$, the ideal number of centroids is 4, and when increasing the number of centroids to 5, the model quality will slightly degrade.

\vspace{2mm}
\textbf{The role of multiple centroids in improving replay buffer:}
In our framework, we leverage centroids obtained from MMOT to improve the quality of the memory buffer. Table \ref{tab:memory_centroids} is an ablation study on the effect of using or not using centroids to select samples for the replay buffer. It can be observed that in general, using centroids to select samples consistently offers better results than random sampling. Regardless of the fact that the sample selection based on centroids is quite simple, the results support our intuition that centroids help effectively and incrementally characterize data in the latent space of the online model, and improve the diversity of the episode memory.

\begin{table}[!ht]
    \centering
    
    \caption{Average Accuracy (\%) when using centroids from MMOT to select samples, and when randomly selecting samples for memory buffer (CIFAR10, $M=1000$)}
    \vspace{-2mm}
    \resizebox{0.4\textwidth}{!}{\begin{tabular}{c cccccc}
        \toprule 
        \multirow{2}{*}{Strategy to select samples for $M$}& \multicolumn{6}{c}{Number of centroids}\\
        \cmidrule(lr){2-7}
        & 1 & 2 & 3 & 4 & 5 & 8 \\
        \midrule
          \rowcolor{gray!10} Based on centroids& \textbf{71.6}  & \textbf{73.8}  & \textbf{75.3} & \textbf{75.9} & \textbf{75.8} & \textbf{74.5}  \\
         Randomly & 68.7 & 71.6 &  73.6& 73.4 & 73.1  & 72.6   \\
         \bottomrule \\
         
    \end{tabular}}
    
    \label{tab:memory_centroids}
    \vspace{-5mm}
\end{table}

\vspace{2.3 mm}
\textbf{The role of centroids in prediction:}
To verify the role of using multiple centroids when making decisions, we present $t$-SNE visualization of the latent space on CIFAR10 (Figure \ref{fig:our_vs_cope}), and MNIST (Figure \ref{fig:ot-gmm}). Figures \ref{fig:our_vs_cope}a,  \ref{fig:our_vs_cope}b, and Figure \ref{fig:ot-gmm} illustrate the effectiveness of our approach when using multiple centroids and only one centroid when predicting. We can see that, in the practical scenario, the features of the classes are usually distributed with multi-modality. Therefore, using adaptive multiple-centroids like ours helps to incrementally characterize that kind of distribution more accurately, thus generally giving better predictions than using only one centroid.

Due to space limitations of the main paper, we present further experimental results in \textbf{\textit{Supplementary \ref{sec:add_exp}.
}}
\section{Conclusion}
\label{sec:conclusion}

This work presents a novel method for Online Class Incremental Learning. Particularly, we introduce an Optimal Transport-driven approach (i.e., MMOT) that can incrementally characterizes data complexity. Based on that, our Dynamic Preservation strategy enhances the model's ability to retain old knowledge, while the MMOT-based testing strategy improves performance. Extensive experiments on benchmark datasets demonstrate the effectiveness of our method, showcasing its potential for improving online learning.

{
    \small
    \bibliographystyle{ieeenat_fullname}
    \bibliography{main}
}


\clearpage
\setcounter{page}{1}
\maketitlesupplementary

\section{Background}
\subsection{Gaussian Mixture Model}
{Formally, a Gaussian Mixture Model (GMM) is a probability distribution composed of several Gaussian components. For a given number of components $K$, it can be expressed as:
\begin{equation*}
\sum_{k=1}^{K} \pi_k \mathcal{N}_k,
\end{equation*}
where each $\mathcal{N}_k$ denotes a Gaussian distribution and the weights satisfy $\sum_{k=1}^{K} \pi_k = 1$. We denote by $GMM_d(K)$ the subset of probability measures in $\mathbb{R}^d$ that can be represented as a Gaussian mixture with at most $K$ components. Note that $K$ can potentially be infinite. The GMM setting is not only a fundamental object in statistical problems but also finds numerous applications, such as image segmentation \citep{farnoush2008image}, anomaly detection \citep{zong2018deep}, keystroke recognition \citep{hosseinzadeh2008gaussian}. \\
Given $n$ i.i.d samples from a distribution $\mathcal{P} \in GMM_d(K)$, the parameters of the GMM representation of $\mathcal{P}$ are typically estimated via maximum likelihood using the Expectation-Maximization (EM) algorithm \citep{dempster1977maximum} with the computational complexity of cubic order (i.e $\mathcal{O}(nKd^3)$). Later, the approach of \citep{pinto2015fast} lowers the computational complexity to $\mathcal{O}(nKd^2)$ by formulating expressions based on precision matrices in place of covariance matrices. \\
The connection between Wasserstein metrics and Gaussian Mixture Models is initially established via the derivation of the optimal transport problem between two GMMs $\sum_{i=1}^{K_1} \pi^{(\mathcal{N})}_i \mathcal{N}_i$ and $\sum_{j=1}^{K_2} \pi^{(\mathcal{P})}_j \mathcal{P}_j$ which can be formulated as the following optimization problem \citep{delon2020wasserstein}:
\begin{equation*}
\min_{\gamma \in \Gamma\left(\pi^{(\mathcal{N})}, \pi^{(\mathcal{P})}\right)} 
\sum_{i=1}^{K_1} \sum_{j=1}^{K_2} \gamma_{i,j} \mathcal{W}_2(\mathcal{N}_i, \mathcal{P}_j).
\end{equation*}
Moreover, \citep{ziesche2023wasserstein} applies Wasserstein Gradient Flows to GMM policy optimization in reinforcement learning. Another family of Wasserstein distances for GMMs is the Orlicz–Wasserstein distance \citep{guha2023excess}. \\

\section{Compare EM and MMOT for OCIL}

\subsection{Computational Complexity and Profiling Surrogates of MOOT}
Let $d$ be the latent dimensionality, $K$ the number of centroids (mixture components) per class, and $B$ the mini-batch size for updating a given class~$c$.
We use diagonal covariances (as in \cref{eq:gmm}), so all Gaussian operations are $\mathcal{O}(d)$ per component.
One MMOT update for class~$c$ (Alg.~2) consists of:
(i) sampling via the reparameterization trick and Gumbel--Softmax,
(ii) evaluating and differentiating the entropic OT dual objective in \cref{eq:dual_OT_GMM} (or its online variant \cref{eq:dual_OT_GMM-batch}),
and (iii) updating the mixture parameters $\{\pi_{k,c}, \mu_{k,c}, \sigma^{2}_{k,c}\}_{k=1}^{K}$.

\begin{itemize}
    \item \textbf{(i) Sampling cost (reparameterization + Gumbel--Softmax).}
For each sample we form $z_k = \mu_{k,c} + \epsilon_k \sigma_{k,c}$ for all $k\!\in\![1, \dots, K]$ and compute relaxed mixture weights
$y_k$ via Gumbel--Softmax, then aggregate $\tilde z = \sum_{k=1}^{K} y_k z_k$.
This requires $\mathcal{O}(BKd)$ floating-point operations and $\mathcal{O}(BK)$ memory for logits and weights.

    \item \textbf{(ii) Dual OT evaluation and gradients.}
The dual objective in \cref{eq:dual_OT_GMM} is an expectation over $P_c$ and $Q_c$.
With a Monte-Carlo estimator using the current mini-batch as samples from $P_c$ and one $\tilde z$ per data point from $Q_c$
(the unbiased single-sample estimator in \cref{eq:dual_OT_GMM-batch}), we evaluate $\phi(f_\theta(x))$, $\tilde{\phi}(\tilde z)$,
and pairwise distances $d(f_\theta(x),\tilde z)$.
Using Mahalanobis with diagonal covariance, distance evaluation is $\mathcal{O}(d)$ per pair.
With one-to-one pairing, the total is $\mathcal{O}(Bd)$; with $S$ negatives per point, it becomes $\mathcal{O}(SBd)$.
Back-propagation through the Kantorovich network $\phi$ (a small MLP) costs $\mathcal{O}(B)$ per pass; repeating $T_\phi$ times per batch (Alg.~2, line 2) yields $\mathcal{O}(T_\phi B)$.

    \item \textbf{(iii) Mixture-parameter updates.}
Gradients for $\{\pi,\mu,\sigma\}$ are linear in $K$ and $d$, i.e.\ $\mathcal{O}(BKd)$, matching the sampling cost.
\end{itemize}

\textbf{Overall complexity.}
A single MMOT update for one class therefore has
\begin{align*}
T_{\text{MMOT}} &= \mathcal{O}\!\left(T_\phi B + BKd + SBd\right), \\
M_{\text{MMOT}} &= \mathcal{O}\!\left(Bd + Kd\right),
\end{align*}
since we never materialize a dense $B{\times}K$ responsibility matrix.
Here $T_\phi$ (typically small) denotes the number of dual-network updates, and $S$ the number of additional negatives (often $S\!\le\!1$).

\subsection{Comparison with EM.}
\paragraph{The classical EM algorithm} (E-step + M-step) for diagonal GMMs evaluates all $K$ component likelihoods for each of the $B$ points and updates sufficient statistics;
both steps are $\mathcal{O}(BKd)$ per iteration, repeated $I_{\text{EM}}\!\gg\!1$ times until convergence:
\[
T_{\text{EM}} = \mathcal{O}\!\left(I_{\text{EM}} BKd\right), 
\qquad
M_{\text{EM}} = \mathcal{O}\!\left(BK + Kd\right),
\]
where the $\mathcal{O}(BK)$ term stores per-point responsibilities across E/M steps.

\paragraph{Centroid-count and drift sensitivity.}
Both EM and MMOT scale linearly in $K$ for diagonal Gaussians.
However, MMOT avoids the inner-loop factor $I_{\text{EM}}$ and the $B{\times}K$ responsibility tensor,
yielding lower memory use and better stability under the continual feature drift characteristic of OCIL.

\underline{\textbf{$\Rightarrow$ Overall}}, we have the complexity summary (per class, per batch) as follow
\[
\begin{array}{l|c|c}
\textbf{Method} & \textbf{Time} & \textbf{Memory} \\
\hline
\text{EM} & \mathcal{O}(I_{\text{EM}}\,BKd) & \mathcal{O}(BK + Kd) \\
\text{MMOT (ours)} & \mathcal{O}(T_\phi B + BKd + SBd) & \mathcal{O}(Bd + Kd) 
\end{array}
\]
When $I_{\text{EM}}$ exceeds a few iterations (as typically required for EM stability),
MMOT becomes asymptotically cheaper in both computation and memory.
Its linear scaling in $K$ matches EM’s, but the constants are smaller thanks to reparameterized sampling and diagonal Mahalanobis distances.
Moreover, MMOT’s single-pass stochastic updates make it better suited to streaming and non-stationary data in OCIL.

\section{Implementation Details}
\label{sec:imp_detail}

\subsection{Datasets:}
As detailed in Section 5 (main text), we employ four datasets to evaluate our method's performance. These original datasets are segmented into various tasks with distinct classes. Below are the specifics regarding the dataset division and task assignments:

\begin{itemize}
    \item \textbf{Tiny-ImageNet} consists of 200 classes, providing 100,000 training samples and 10,000 test samples, with images sized at 64 × 64 pixels. It is divided into 100 non-overlapping tasks, each containing two classes.
    \vspace{2mm}
    \item \textbf{CIFAR100} includes 100 classes, offering 50,000 training samples and 10,000 test samples, also at 32 × 32 pixels. This dataset is split into 10 separate tasks, with 10 classes per task.
    \vspace{2mm}
    \item \textbf{CIFAR10} contains 10 classes, with 50,000 training samples and 10,000 test samples, all sized at 32 × 32 pixels. For our experiments, it is divided into five non-overlapping tasks, each featuring two classes. 
    \vspace{2mm}
    \item \textbf{MNIST} consisting of 60,000 training samples and 10,000 test samples of handwritten digits (0 through 9). Each image is 28 × 28 pixels in size. For our experiment, it is splitted into 5 disjoint subsets, corresponding to 5 tasks, each of which consists of 2 classes.
\end{itemize}
\vspace{2mm}
For the streaming input data, we set the
batch size to 10, and for the samples drawn from the buffer,
the batch size is set to 64. We also employ data augmentation strategy for our method and baselines, as describe in \citep{wang2024improving}.




\subsection{Model Architectures:} 
For the experiments on {MNIST} dataset, we use a simple MLP neural network with 2 hidden layers of 400 units. While a slim version of ResNet-18 will be used to evaluate on three remaining datasets as commonly used in other recent state-of-the-art baselines \citep{onpro, szatkowski2025improving}.

\subsection{Evaluation and metrics}


We used the following metrics to evaluate:
\begin{itemize}
    \item \textbf{Average accuracy} ($\mathcal{A}_T$): Averaged test accuracy of all tasks after completing learning T task. 
    $$\mathcal{A}_T = \frac{1}{T}\sum_{i=1}^T a_{i},$$
    where $a_{i}$ is the accuracy at the end of the $i^{th}$ task. 
    
    \item \textbf{Average forgetting} ($\mathcal{F}_T$): The averaged gap between the highest recorded and final task accuracy at the end of continual learning on $T$ tasks.
    $$ \mathcal{F}_T = \frac{1}{T-1}\sum_{i=1}^{T-1}f_{i},$$
    where $f_{i}$ is forgetting of task $i^{th}$ after learning task $T$.
\end{itemize}

The experiments were performed over multiple runs, each with varying sequences of incoming classes. We provide the mean and standard deviation to illustrate the robustness of our results across different class orderings and random seeds.

\section{Additional Experiments}
\label{sec:add_exp}

\subsection{Performance comparison on MNIST}


Table \ref{tab:bal_compare_forget_2} further provides a comparison between our method (OTC) and the three strongest baselines GSA, MOSE, and BiC+AC on dataset MNIST. The results show that our method consistently outperforms all these baselines in both average accuracy (up to 2.4 \%) and forgetting (up to 1.6 \%). 





\begin{table}[!ht]
    \centering
    
    \resizebox{0.45\textwidth}{!}{\begin{tabular}{c cc|cc}
        \toprule 
        \multirow{2}{*}{Method} & \multicolumn{2}{c}{M = 0.5k} & \multicolumn{2}{c}{M = 1.5k}\\
        \cmidrule(lr){2-3} \cmidrule(lr){4-5} 
        & Acc $\uparrow$ & Forgetting $\downarrow$ & Acc $\uparrow$ & Forgetting $\downarrow$  \\
        \midrule
        \midrule
            GSA &  92.7 ± 0.3 & 3.3 ± 0.2 & 96.8 ± 0.1 & 1.8 ± 0.2 \\ 
            MOSE & 91.2 ± 0.4 & 4.0 ± 0.5 & 96.9 ± 0.2 & 1.2 ± 0.3 \\
            BiC+AC & 93.0 ± 0.5 & 2.8 ± 0.6 & 97.2 ± 0.4 & 1.2 ± 0.3\\
            
        \midrule
            OTC & \textbf{93.6 ± 0.3} & \textbf{2.4 ± 0.4} & \textbf{97.7 ± 0.4} & \textbf{1.1 ± 0.4} \\
         \midrule \\
          
         
    \end{tabular}}
    \vspace{-5mm}
    \caption{Evaluations on MNIST.}
    \label{tab:bal_compare_forget_2}
\end{table}

\subsection{Offline setting}
\begin{table}[!ht]
    \centering
    \vspace{-2mm}
    \resizebox{0.465\textwidth}{!}{\begin{tabular}{l rrr|rrr}
        \toprule 
        Method & M = 200 & M = 500 & M = 5120 & M = 200 & M = 500 & M = 5120 \\
        \midrule

         DER++ &64.88 &72.70 &85.24 &18.66 &28.70 &51.20 \\
         GeoDL &49.20 &61.83 &85.91 &13.38 &23.06 &54.57 \\
         Co2L &65.57 &74.26 &84.27 &18.85 &24.45 &46.18 \\
         \midrule
         OTC (Ours) & \textbf{67.05} & \textbf{76.45} & \textbf{87.03} & \textbf{25.22} & \textbf{33.18} & \textbf{57.32}\\
         \midrule
         & \multicolumn{3}{c}{a) CIFAR10} & \multicolumn{3}{c}{b) CIFAR100}\\
          
         
    \end{tabular}}
    \vspace{-3mm}
    \caption{Average Accuracy ($\uparrow$) in the offline setting of CIL, M: buffer size.}
    \label{tab:offline}
\end{table}
Table \ref{tab:offline} examines the behavior of OTC in the offline setting of Class Incremental Learning when compared with some typical offline methods, including DER++ \citep{DBLP:conf/nips/BuzzegaBPAC20}, GeoDL \citep{DBLP:conf/cvpr/SimonKH21}, and Co2L \citep{Cha_2021_ICCV}. The results show that OTC demonstrates its superiority across all considered cases, with the most significant gap can be more than 6\% compared to the strongest considered baseline. This demonstrates our effective application of our method in both online and offline setting of Class Incremental Learning problem.

\end{document}